\documentclass[runningheads]{llncs}
\usepackage{graphicx}
\usepackage{comment}
\usepackage{amsmath,amssymb} 
\usepackage{color}
\usepackage[colorlinks,linkcolor=blue]{hyperref}
\usepackage{multirow}

\begin{document}
\pagestyle{headings}
\mainmatter
\def\ECCVSubNumber{6037}  

\title{Pose Augmentation: Class-agnostic Object Pose Transformation for Object Recognition} 

\titlerunning{Class-agnostic Object Pose Transformation}
%
\author{Yunhao Ge\inst{1}\orcidID{0000-0002-8110-9280} \and
Jiaping Zhao\inst{2} \and \\
Laurent Itti\inst{1}\orcidID{0000-0002-0168-2977}}
\authorrunning{Y. Ge et al.}
%
\institute{University of Southern California, Los Angeles, USA \email{\{yunhaoge,itti\}@usc.edu}\\ 
\and
Google Research, Los Angeles, USA \email{jiapingz@google.com}}
\maketitle
\begin{abstract}
Object pose increases intraclass object variance which makes object recognition from 2D images harder. To render a classifier robust to pose variations, most deep neural networks try to eliminate the influence of pose by using large datasets with many poses for each class. Here, we propose a different approach: a class-agnostic object pose transformation network (OPT-Net) can transform an image along 3D yaw and pitch axes to synthesize additional poses continuously. Synthesized images lead to better training of an object classifier. We design a novel eliminate-add structure to explicitly disentangle pose from object identity: first ‘eliminate’ pose information of the input image and then ‘add’ target pose information (regularized as continuous variables) to synthesize any target pose. We trained OPT-Net on images of toy vehicles shot on a turntable from the iLab-20M dataset. After training on unbalanced discrete poses (5 classes with 6 poses per object instance, plus 5 classes with only 2 poses), we show that OPT-Net can synthesize balanced continuous new poses along yaw and pitch axes with high quality. Training a ResNet-18 classifier with original plus synthesized poses improves mAP accuracy by 9\% over training on original poses only. Further, the pre-trained OPT-Net can generalize to new object classes, which we demonstrate on both iLab-20M and RGB-D. We also show that the learned features can generalize to ImageNet. (The code is released at this \href{https://github.com/gyhandy/Pose-Augmentation}{URL}) 
\keywords{pose transform, data augmentation, disentangled representation learning, object recognition, GANs}
\end{abstract}
\begin{figure}
\centering
\includegraphics[height=2in]{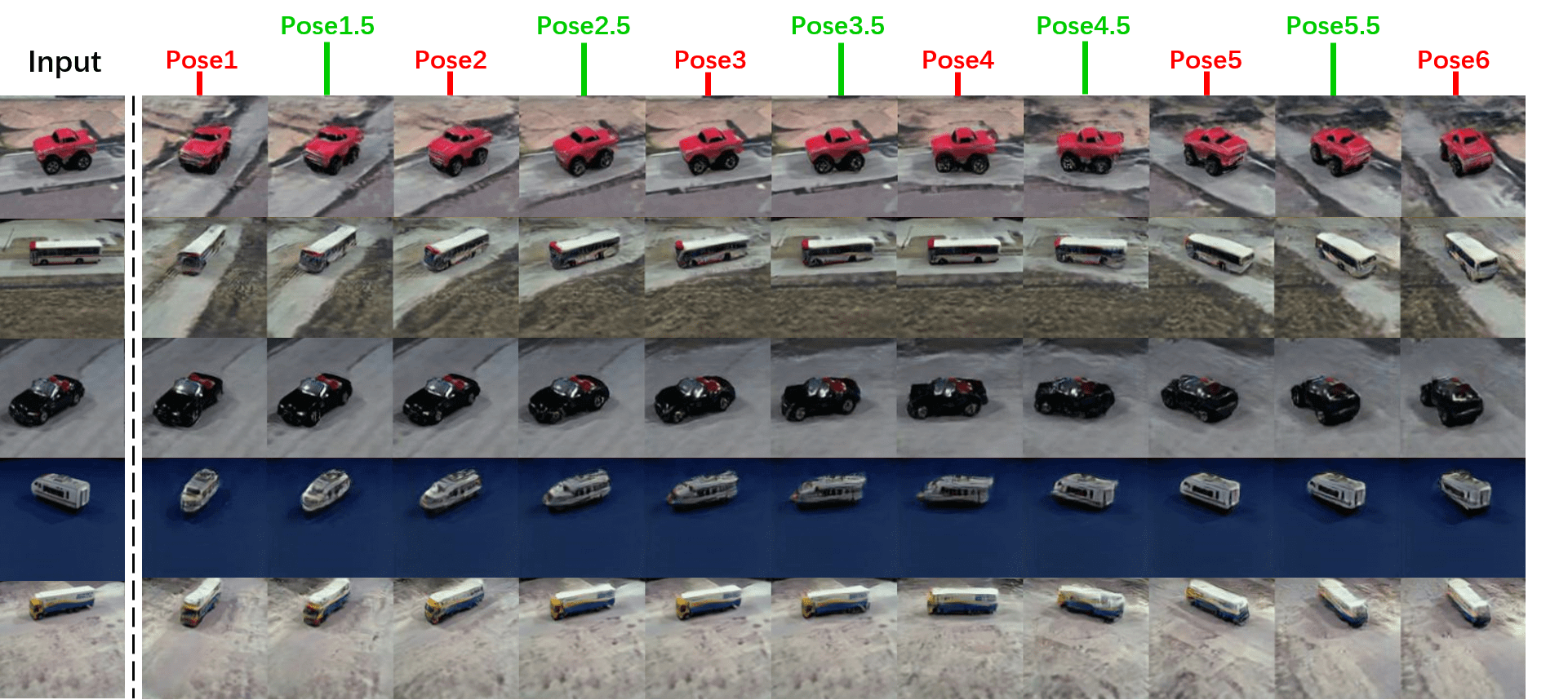}
\caption{Object pose transformation with OPT-Net. The first column shows input images from the test dataset, and the remaining columns show target pose images transformed by OPT-Net.  Integer poses (1, 2, 3, 4, 5, 6 in red) are defined in the training dataset, while decimal poses (1.5, 2.5, 3.5, 4.5, 5.5 in green) are new poses, which shows OPT-Net can achieve continuous pose transformation. }
\label{fig1}
\end{figure}
\section{Introduction and related work}

In object recognition from 2D images, object pose has a significant influence on performance. An image depends on geometry (shape), photometry (illumination and material properties of objects) and dynamics (as objects move) of the scene. Thus, every image is a mixture of instance-specific information and nuisance factors ~\cite{ma2012invitation}, such as 3D viewpoint, illumination, occlusions, shadows, etc. Nuisance factors often depend on the task itself. Specifically, in object recognition from 2D images, we care for instance-specific information like shape, while the dynamics of pose is a nuisance that often degrades classification accuracy ~\cite{ma2012invitation}. 

Deep convolution neural networks (CNNs) have achieved great success in object recognition \cite{krizhevsky2012imagenet,simonyan2014very,he2016deep,szegedy2017inception,huang2017densely} and many other tasks, such as object detection \cite{girshick2014rich,ren2015faster,redmon2016you,fang2017rmpe}, image segmentation  \cite{ronneberger2015u,milletari2016v,he2017mask}, etc. Most research tries to discount pose, by eliminating pose information or improving pose robustness of a classifier. Typical CNN architectures, such as LeNet \cite{lecun2015lenet} AlexNet \cite{krizhevsky2012imagenet} and VGG \cite{simonyan2014very} use convolution layers and pooling layers to make the high-level feature representations invariant to object pose over some limited range \cite{zhao2017learning}. In contrast, recent results have shown that explicitly modeling pose information can help an object recognition task \cite{zhao2017learning,bengio2013representation,wohlhart2015learning,bakry2014untangling}. Some approaches use multi-task learning where pose information can be an auxiliary task or regularization to improve the main object recognition task \cite{zhang2014facial,zhang2014improving,huang2013multi,su2015render}. These neural networks have the potential to disentangle content from their instantiation attributes \cite{ranzato2007unsupervised,zhao2015stacked,goroshin2015learning}. Training on multiple views of the object can improve recognition accuracy \cite{su2015multi}. A common method is collecting all poses of the object and creating a pose-balanced dataset, with the hope that pose variations will average out. However, collecting pose-balanced datasets is hard and expensive. One notable such dataset is iLab-20M which comprises 22 million images of 704 toy vehicles captured by 11 cameras while rotating on a turntable \cite{borji2016ilab}. Here, we use a subset of this data to learn about pose transformations, then transferring this knowledge to new datasets (RGB-D \cite{lai2011large}, ImageNet \cite{deng2009imagenet}).

2D  images can be seen as samples of 3D poses along yaw and pitch axes (Fig.2(a)). We want our OPT-Net to imitate the 3D pose transformation along these two axes. Thus given any single pose image, we can 'rotate' the object along yaw and pitch axes to any target pose. Instead of directly training a transformation model to continuously 'rotate' images, we start with a discrete transform, which is easier to constrain. Then we can make the pose representation continuous and regularize the continuous transform process.
Here, we use sampled discrete poses along yaw and pitch as our predefined poses (Fig.2(b), 6 poses along the yaw axis and 3 poses along pitch axis). We treat different object poses as different domains so that discrete pose transformation can be seen as an image-to-image translation task, where a generative model can be used to synthesize any target pose given any input pose. Recently, Generative Adversarial Networks (GAN) \cite{goodfellow2014generative} have shown a significant advantage in transforming images from one modality into another modality \cite{mirza2014conditional,isola2017image,sangkloy2017scribbler,zhu2017unpaired,kim2017learning}. GANs show great performance in various tasks, such as style transfer \cite{chang2018pairedcyclegan,karras2019style}, domain adaptation \cite{hoffman2017cycada,tzeng2017adversarial}, etc. However, there is a high cost in our task, because we should train specific GANs for all pairs of poses \cite{bhattacharjee2018posix}. StarGAN \cite{choi2018stargan} and CollaGAN \cite{lee2019collagan} proposed a method for multi-domain mapping with one generator and showed great results in appearance changes such as hair color, age, and emotion transform. However, pose transform creates a large, nonlinear spatial change between input and output images. The traditional structure of the generators (Unet \cite{ronneberger2015u}, Vnet \cite{milletari2016v}) has few shared structures which satisfy all randomly paired pose transformation. It makes StarGAN training hard to converge (see Exp 4.1). 

Learning a better representation could also reduce variance due to pose. \cite{zhu2014multi} tried to learn better representation features to disentangle identity  rotation and view features. InfoGAN \cite{chen2016infogan} learns disentangled representations in an unsupervised manner. \cite{kan2016multi} seeks a view-invariant representation shared by views.

To combine the idea of better representation and multi-domain image transformation, we propose a class-agnostic object pose transformation neural network (OPT-Net), which first transforms the input image into a canonical space with pose-invariant representation and then transform it to the target domain. We design a novel eliminate-add structure of the OPT-Net and explicitly disentangle pose from object identity:  OPT-Net first ‘eliminates’ the pose information of the input image and then ‘adds’ target pose information to synthesize any target pose. Convolutional regularization is first used to implicitly regularize the representation to keep only the key identification information that may be useful to any target pose. Then, our proposed pose-eliminate module can explicitly eliminate the pose information contained in the canonical representation by adversarial learning. We also add a discriminator leveraging pose classification and image quality classification to supervise the optimization of transforming.


Overall our contributions are multifold: 
(1) developed OPT-Net, a novel class-agnostic object pose transformation network with an eliminate-add structure generator that learns the class-agnostic transformation among object poses by turning the input into a pose-invariant canonical representation.
(2) design a continuous representation of 3D object pose and achieve continuous pose transforming in 3D, which can be learned from limited discrete sampled poses and adversarial regularization. 
(3)	demonstrated the generative OPT-Net significantly boosts the performance of discriminative object recognition models. 
(4)	showed OPT-Net learns class-agnostic pose transformations, generalizes to out-of-class categories and transfers well to other datasets like RGB-D and ImageNet.

\begin{figure}[t]
\centering
\includegraphics[height=2cm]{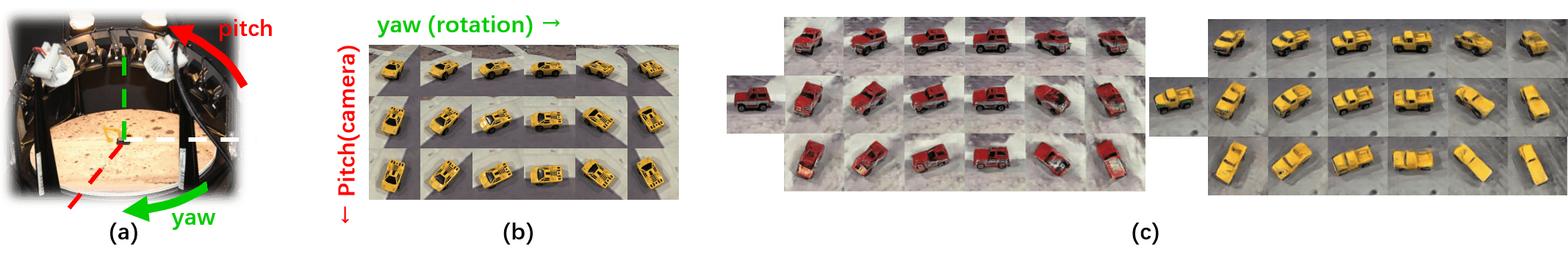}
\caption{(a) Discrete predefined pose images sample. (b) Predefined sample poses and pose change along pitch and yaw axes. (c) Given any pose (1st and 8th columns), OPT-Net can transform it along pitch and yaw axes to target poses (remaining columns)}
\end{figure}

\section{Object Pose Transforming Network}

As shown in Fig. 3, the proposed OPT-Net has an eliminate-add structure generator, a discriminator and a pose-eliminate module.

\subsection{Eliminate-add structure of the generator}

\begin{figure}[b!]
\centering
\includegraphics[width=5in]{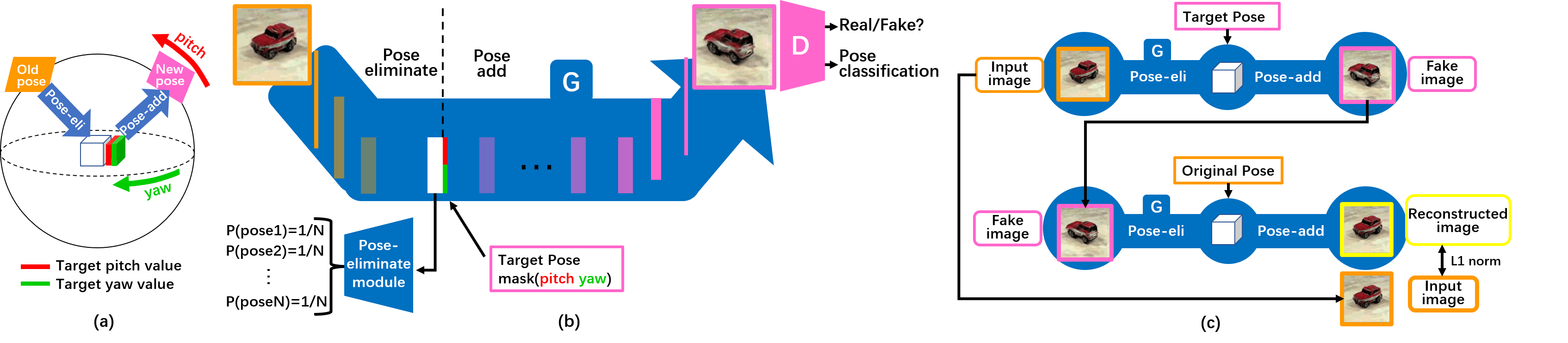}
\caption{Flow of OPT-Net, consisting of three modules: eliminate-add structure generator $G$, discriminator $D$, and pose-eliminate module. (a) Pose transformation sketch (b) Origin to target pose transformation. In the pose ‘eliminate’ part, $G$ takes in the original pose image and first uses both implicit regularization and the explicit pose-eliminate module to eliminate pose information of the input, yielding a pose-invariant canonical representation. Then, in the pose ‘add’ part, the representation features are concatenated with a target pose mask and the target pose image is synthesized. $D$ learns to distinguish between real and fake images and to classify real images to their correct pose. (c) Training OPT-Net: $G$ first maps the original pose image to target pose and synthesizes a fake image, then $G$ tries to reconstruct the original pose image from the fake image given the original pose information.}
\end{figure}

The generator ($G$) of OPT-Net transforms an input object pose image x into a target object pose y conditioned on the target pose label $c$, $G(x,c)\to$y$ $. Different from the hair color, gender and age transform, which have more appearance transfer with smaller shape changes, object pose transformation creates large shape differences. Our eliminate-add structure generator (Fig.3 (a)) first turns the input pose image into a pose-invariant canonical representation by ‘eliminating’ pose information, and then ‘adds’ target pose information to turn the representation into the target pose. As shown in Fig. 3(b), given an input image, we randomly select the target pose domain. We do not input target pose along with the input image. Instead, in the ‘eliminate’ part, the first several convolution layers with stride $s>2$ are used to implicitly regularize the preserved representation features. This implicit regularization makes the representation features contain only key information for the transformation (appearance, color, shape), and eliminates useless information which may hinder transformation (pose). At the same time (Fig. 3(b)), the ‘pose-eliminate module’ ($P_{elim}$) explicitly forces the representation to contain as little pose information as possible, by predicting equal probability for every pose. After both implicit and explicit elimination of pose information, the input image is turned to a pose-invariant canonical representation space. We then ‘add’ the target pose information by concatenating it with the representation feature map. The remaining layers in the generative model transform the concatenated features into the target pose image. This eliminate-add structure is shared and can be used for any pose transformation. This shared structure makes the generator easy to converge.
To control the translation direction, as shown in Fig. 3(b), we use an auxiliary classifier as discriminator $D$ to guide the image quality and pose transform. Given one image, the discriminator has two outputs, the probability that the input image is real, which represents the quality of the synthesized image, and the output pose, which should match the desired target pose, $D:x\to\{ D_{src}(x), D_{cls} (x)\} $

\subsection{Pose-eliminate module}
The pose-eliminate module ($P_{elim}$) takes the preserved representation feature $x_r$ as input and outputs pose classification $\{ P_{elim}(x_r) \}$. $P_{elim}$ can be treated as a discriminator which forms an adversarial learning framework with the ‘eliminate’ part of the generator ($G_{elim}$).
The canonical representation features of real images with pose labels are used to train $P_{elim}$. We use Cross-Entropy loss to make $P_{elim}$ predict the correct pose from the pose-invariant feature after $G_{elim}$. Different from  traditional adversarial training, when using $P_{elim}$ to train $G_{elim}$, we want the generator to eliminate all pose information in the pose-invariant feature, which makes $P_{elim}$ produce equal output probability for every pose. We use the uniform probability ($1/N$) as the ground truth label to compute the pose-eliminate loss, which is used to optimize the $G_{elim}$.

\subsection{Continuous pose transforming training}
We design a 2-dimension linear space to represent  pitch and yaw values, in which we could interpolate and achieve continuous pose representation (Fig.1). The yaw and pitch values can be duplicated as a matrix with same $h$ and $w$ dimension as the canonical representation features and N (totally 6, 3 for yaw and 3 for pitch) channel dimension, which is easy to be concatenated and can be adjusted depending on the canonical features channel. We start the training on discrete sampled poses (which can be represented as integer in linear space). After the network has converged, we randomly sample decimal poses as target poses and use a style consistency loss to regularize the synthesized images, which keeps pose representation consistent along yaw and pitch axes. 


\subsection{Loss Function}
Our goal is to train a generator $G$ that learns object pose transformations along yaw and pitch axes. The overall loss is formed by adversarial loss, domain classification loss, reconstruction loss, pose-eliminate loss and style consistency loss.


\noindent{\bf Adversarial Loss.} The adversarial loss is used to make the synthesized image indistinguishable from real images.
\begin{equation}\label{Eq.1}
	L_{adv}=E_{x}[logD_{src}(x)] + E_{x,c}[log(1-D_{src}(G(x,c)))]
\end{equation}
$ D_{src}(x) $ represent the probability that input x belongs to the real images given by $D$. The generator $G$ tries to minimize the loss, while the discriminator $D$ tries to maximize it. 


\noindent{\bf Pose Classification Loss.}
The pose classification loss is used to guide the pose transformation which makes the synthesized image $y$ belong to the target pose $c$. This pose classification loss is used to optimize both $D$ and $G$. The pose classification loss of $D$ is defined as
\begin{equation}\label{Eq.2}
	L_{cls}^{r}=E_{x,c'}[-logD_{cls}(c'|x)] 
\end{equation}
The loss for $D$ is similar to a traditional Cross-Entropy loss for classification, where$ D_{cls}(c'|x) $ means the predicted probability of real image x belongs to the ground truth pose label $c' $ . The pose classification loss of $G$ is defined as
\begin{equation}\label{Eq.3}
	L_{cls}^{f}=E_{x,c}[-logD_{cls}(c|G(x,c))] 
\end{equation}
$G$ tries to minimize this loss to make the synthesized fake image $G(x,c) $ be classified as the target pose $c$.

\noindent{\bf Reconstruction Loss.}
To make the synthesized image preserve the content information and change only the object pose, as shown in fig3(c), we use the cycle consistency loss \cite{zhu2017unpaired} to optimize $G$.
\begin{equation}\label{Eq.4}
	L_{rec}=E_{x,c,c'}[\|x-G(G(x,c), c')\|_{1}] 
\end{equation} 
where $G$ can reconstruct the original image $x$ by transforming the synthesized fake target pose image $G(x,c) $ back to the original pose $c'$ . $L1$ norm is used as reconstruction loss.

\noindent{\bf Pose-eliminate Loss.}
In the eliminate-add structure of $G$, to eliminate the pose information in preserved canonical representation features, we designed pose-eliminate loss to optimize the pose eliminate module ( $P_{elim}$) and the eliminate part of $G$,($G_{elim}$ ). The pose eliminate loss is
\begin{equation}\label{Eq.5}
	L_{pose}^{P}=E_{x,c'}[-logP_{elim}(c'|G_{elim}(x))] 
\end{equation} 
where $P_{elim}(c'|G_{elim}(x)) $  means the predicted probability of the canonical representation features   of a real image belongs to the ground truth pose label $c' $ . 
The pose eliminate loss for $G_{elim}$  is defined as
\begin{equation}\label{Eq.6}
	L_{pose}^{G}=-E_{x}{\sum_{c_{i}=1}^{N}1/N\cdot log(P_{elim}(c_{i}|G_{elim}(x)))} 
\end{equation} 
where $N$ is the number of pose classes we defined, $c_{i} $ represent the pose label, $c_{i}\in [0, N) $ , $P_{elim}(c_{i}|G_{elim}(x))$ represent the probability of the synthesized canonical representation belongs to the $c_{i} $  pose. In ideal situations, the $P_{elim}$  can hardly predict the correct pose from canonical representation features and output equal probability for every pose, which means the pose information is eliminated in preserved canonical features. We use equal prediction of every pose to optimize $G_{elim}$ instead of minimizing the pose classification accuracy of  to avoid a ‘cheated optimize’ that $P_{elim}$  tries to predict all input to a fixed pose class. 

\noindent{\bf Style consistency Loss.}
After the converge of the previous loss, we randomly sample decimal target pose instead of all integers to make continuous pose transforming, the style consistency loss can regularize the synthesized images. The equation of style consistency loss is same as adversarial loss above, but the target pose is randomly sampled decimal value along yaw and pitch axes.

\noindent{\bf Full Loss Function.}
Finally, we optimize:
\begin{equation}\label{Eq.8}
	L_{G}=L_{adv} + \lambda_{cls}L_{cls}^{f} + \lambda_{rec}L_{rec} + \lambda_{pose}L_{pose}^{G} 
\end{equation} 
\begin{equation}\label{Eq.9}
	L_{D}=-L_{adv} + \lambda_{cls}L_{cls}^{r} 
\end{equation} 
\begin{equation}\label{Eq.10}
	L_{P_{elim}}=L_{pose}^{P} 
\end{equation} 
where $ \lambda_{cls}$, $ \lambda_{rec} $  and $ \lambda_{pose} $  are hyper-parameters that control the relative importance of classification, reconstruction, and pose-eliminate losses.
\section{Experimental Methods}
\subsection{Datasets}

\noindent{\bf iLab-20M dataset \cite{borji2016ilab}.} The iLab-20M dataset is a controlled, parametric dataset collected by shooting images of toy vehicles placed on a turntable using 11 cameras at different viewing points. There are in total 15 object categories with each object having 25~160 instances. Each object instance was shot on more than 14 backgrounds (printed satellite images), in a relevant context (e.g., cars on roads, trains on rail tracks, boats on water). In total, 1,320 images were captured for each instance and background combinations: 11 azimuth angles (from the 11 cameras), 8 turntable rotation angles, 5 lighting conditions, and 3 focus values (-3, 0, and +3 from the default focus value of each camera). The complete dataset consists of 704 object instances, with 1,320 images per object-instance/background combination, almost 22M images (18 times of ImageNet).

\noindent{\bf RGB-D dataset.} The RGB-D Object Dataset consists of 300 common household objects organized into 51 categories. This dataset was recorded using a Kinect style 3D camera. Each object was placed on a turntable and video sequences were captured for one whole rotation. For each object, there are 3 video sequences, each recorded with the camera mounted at a different height so that the object is shot from different viewpoints.
\subsection{Network Implementation}
OPT-Net consists of two parts, pose ‘eliminate’, ( including $G_{elim}$ and $P_{elim}$) and pose ‘add’, (including $G_{add}$ and $D$). As shown in Fig.~3(b), $G_{elim}$ first has 3 convolution layers, 2 of them with stride size of 2 to down-sample the input image. Then, 3 Residual blocks \cite{he2016deep} form the backbone of $G_{elim}$. The output $G_{elim}(x)$  is the pose-invariant canonical representation feature. The canonical feature is copied to different streams, one concatenates with the target pose mask, forming the input of $G_{add}$ to synthesize the target pose image. The other one is treated as the input of $P_{elim}$ to predict the pose class. $G_{add}$ uses first layer merge the target pose information, then has 5 Residual blocks as a backbone and ends with 3 convolution layers (2 of them perform up-sampling) to transform the canonical representation features to a target pose image, given a target pose information mask. For discriminator $D$, we adopt the PatchGAN \cite{isola2017image} network.

$P_{elim}$ has a traditional classification network structure, which has the first 3 convolution layers with stride size of 2 to down-sample the input features, followed with 1 Residual block and another 3 down-sampling convolution layers. In the end, the output layer turns the feature to a N-dimensional (N poses) vector and we use Softmax to obtain the prediction of pose class.

We use Wasserstein GAN objective with a gradient penalty \cite{arjovsky2017wasserstein} to stabilize the training process.
We adjust the $ \lambda_{pose} $ during training the generator, at the beginning epochs of training, improving the value of $ \lambda_{pose} $  can accelerate the convergence of generator, which makes the synthesized fake pose image have meaningful corresponding spacial structure. We gradually reduce the value of $ \lambda_{pose} $. At the last ending part of the training,  $ \lambda_{pose} $ can be very small to make the optimization concentrate on improving the image quality.
(More network architecture and training details are in supplementary materials )
\section{Experiments and Results}
We have five main experiments: in Section 4.1 on object pose transformation task, we compare  OPT-Net with baseline  StarGAN \cite{choi2018stargan} by quantitatively and qualitatively comparing the synthesized object pose image quality. In Section 4.2, we use the OPT-Net as a generative model to help the training of a discriminative model for object recognition, by synthesizing  missing poses and balancing a pose bias in the training dataset. In Section 4.3, we further show the class-agnostic transformation property of OPT-Net by generalizing the pretrained OPT-Net to new datasets. In Section 4.4, we study the influence of object pose information for objects which are mainly distinguishable by shape, as opposed to other features like color. Finally, in Section 4.5, we further demonstrate how the learned pose features in OPT-Net and object recognition model with the iLab-20M dataset can  generalize to other datasets like ImageNet.
\subsection{Object Pose Transformation Experiments}
Because the baseline models can only do discrete pose transform, we fix the pitch value and use 6 different yaw viewpoints among the 88 different views of iLab-20M as our predefined pose to implement our OPT-Net. As is shown in Fig. 2, the selected 6 viewpoints have big spatial variance which can better represent the general object pose transformation task. In training set, each pose has nearly 26k images with 10 vehicle classes (Table 2). Each class contains 20$\sim$80 different instances. The test set has the same 10 vehicle categories, but different instances than the training set. Both training and test datasets are 256x256 RGB images. The training dataset is used to train our OPT-Net and the baseline models, StarGAN. Our OPT-Net has one generator, one discriminator and one pose-eliminate module; StarGAN has one generator and one discriminator. 

\begin{figure}[t]
\centering
\includegraphics[width=\linewidth]{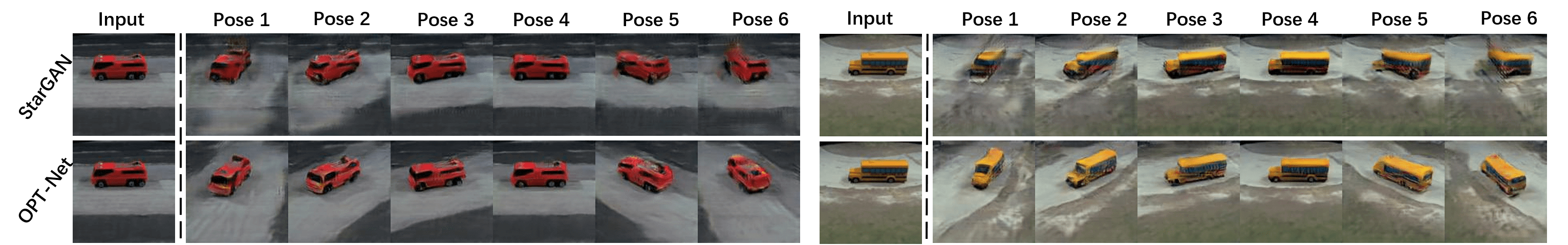}
\caption{Object pose transform comparison for StarGAN and OPT-Net.}
\end{figure}

\textbf{Qualitative evaluation}. The experiment results are shown in Fig. 4. Compared with StarGAN, which struggles with large pose variations, the synthesized target pose images by OPT-Net are high quality with enough details. One possible reason is that eliminate-add structure decrease the conflicts between different directions on pose transformation. Fig.1 shows more results of OPT-Net. 

\textbf{Quantitative evaluation}. Real target pose images of input are used as ground truth. To reduce background influence, we segment the foreground vehicle with the Graph-Based Image Segmentation method and only compute mean squared error (MSE) and peak signal to noise ratio (PSNR) of foreground between the synthesized image and ground truth (Table 1). 
The result is the mean MSE and PSNR computed by 200 different instances, the MSE and PSNR for each instance is the average of 6 synthesized fake pose images. Table 1 shows that the quality of synthesized images by OPT-Net is better than StarGAN.

\begin{table}

\begin{center}
\caption{Average Mean squared error (MSE; lower is better) and peak-signal-to-noise ratio (PSNR; higher is better) for different methods}
\begin{tabular}{llll}
\hline\noalign{\smallskip}
  &   StarGAN \ \ \ & OPT-Net \\
  \noalign{\smallskip}
 \hline
 Mean MSE &  502.51 & \bf{374.76}\\
 Mean PSNR \ \ \  &  21.95 &  \bf{23.04}\\
\hline
\end{tabular}
\end{center}

\end{table}

\subsection{Object Recognition Experiment}
We design an object recognition experiment to explore the performance of OPT-Net as a generative model to help the training of a discriminative model. Two different training datasets are tailored from  iLab-20M, pose-unbalanced (P-UB) and pose-balanced (P-B). In P-UB (Table~2), 5 classes of vehicles (boat, car, semi, tank, and van) have all 6 pose images (same poses as 4.1), while the other 5 classes  (bus, military car, monster, pickup, and train)  have only two poses (pose2 and pose5), which has significant pose bias. In P-B, each category among 10 classes of vehicles has all 6 pose images (no pose bias). The test dataset is a pose-balanced dataset which contains different instances of the 10 classes of vehicles that were not in either training dataset (P-UB and P-B). The classification neural network  we used is Resnet-18 \cite{he2016deep} (no  pre-training).

\begin{table}
\begin{center}
\caption{Poses used in the pose-unbalanced (P-UB) training dataset to train  OPT-Net }
\begin{tabular}{lllllll}
\hline\noalign{\smallskip}
 &  Pose1\ \ & Pose2\ \ & Pose3\ \ & Pose4\ \ & Pose5\ \ & Pose6 \\
\noalign{\smallskip}
\hline
boat & \checkmark & \checkmark & \checkmark & \checkmark & \checkmark & \checkmark\\
bus &  & \checkmark &  &  & \checkmark & \\
car & \checkmark & \checkmark & \checkmark & \checkmark & \checkmark & \checkmark\\
mil &  & \checkmark &  &  & \checkmark & \\
monster &  & \checkmark &  &  & \checkmark & \\
pickup &  & \checkmark &  &  & \checkmark & \\
semi & \checkmark & \checkmark & \checkmark & \checkmark & \checkmark & \checkmark\\
tank & \checkmark & \checkmark & \checkmark & \checkmark & \checkmark & \checkmark\\
train &  & \checkmark &  &  & \checkmark & \\
van & \checkmark & \checkmark & \checkmark & \checkmark & \checkmark & \checkmark\\
\hline
\end{tabular}
\end{center}
\end{table}

We first train the classification model on P-UB and P-B, calculating the test accuracy of each class of vehicles on the test dataset. To evaluate the performance of OPT-Net, we first train it on P-UB  to learn the object transformation ability. After training, for each category in P-UB which have only pose2 and pose5 (bus, military car, monster, pickup, and train), we use the trained OPT-Net to synthesize the missing 4 poses (pose1, pose3, pose4, pose6). We combine the synthesized images with P-UB and form a synthesized-pose-balanced (S-P-B) training dataset. To show  continuous transforms, we also interpolate  pose values and synthesize 5 new poses beyond the predefined ones, and form a synthesized-additional-pose-balanced (SA-P-B) training dataset. S-P-B and SA-P-B were used to train the same resnet-18 classification model from scratch and to calculate test accuracy of each class of vehicles in the test dataset. We also use common data augmentation methods (random crop, horizontal flip, scale resize, etc) to augment the P-UB dataset to the same number of images as P-B, called A-P-UB (Table~3).

The test accuracy of each class is shown in Table 4. From P-UB to S-P-B, the overall accuracy improved from 52.26\% to 59.15\%, which shows the synthesized missing pose images by OPT-Net can improve the performance of object recognition. It is also shown that OPT-Net, as a generative model, can help the discriminative model. Specifically, the vacant pose categories show significant improvement in accuracy: military improved by 11.68\%, monster improved by 14.97\%, pickup and train improved by 8.74\% and 16.12\% respectively. The comparison of S-P-B and A-P-UB shows that synthesized images by OPT-Net are better than traditional augmented images in helping object recognition. Because of the continuous pose transformation ability, our OPT-Net can synthesize additional poses different from the 6 poses in P-B. With these additional poses, SA-P-B (61.23\%) performs even better than the P-B (59.20\%), achieve 9\% improvement compared with P-UB.

\begin{table}[t]
\begin{center}
\caption{Different training and testing datasets for object recognition}

\begin{tabular}{lllllll}
\hline\noalign{\smallskip}
Dataset\ \ \ & P-UB\ \ \ & P-B\ \ \ \ \ \ & S-P-B\ \ \ \ \ \ \ \ \ \  & SA-P-B\ \ \ \ \ \ \  & A-P-UB\ \ \ \ \ \ \  & Test  \\
\noalign{\smallskip}
 \hline
Source & real & real & synthesized & synthesized & augmented & real \\
Size & 25166 & 37423 & 37423 & 66041 & 37423 & 4137 \\
 \hline
\end{tabular}
\end{center}
\end{table}

\begin{table}
\begin{center}
\caption{Testing object recognition accuracy (\%) of each class after trained on different training dataset. Comparing S-P-B and SA-P-B with P-UB shows how much classification improves thanks to adding synthesized images for missing poses in the training set,  reaching or surpassing the level of when all real poses are available (P-B). Our synthesized poses yield better learning than traditional data augmentation (A-P-UB)}
\begin{tabular}{llllll}
\hline\noalign{\smallskip}
Category\ \ \ & P-UB\ \ \ \ & P-B\ \ \ \ \ & S-P-B\ \ \ \ & SA-P-B\ \ & A-P-UB\\
\noalign{\smallskip}
 \hline
boat & 54.0 & 61.6 & 65.4 & 57.7 & 51.3\\
\textbf{bus} & 35.2 & 42.5 & 38.1 & 47.8 & 37.2 \\
car & 85.1 & 76.3 & 79.8 & 64.0 & 78.9 \\
\textbf{mil} & 73.8 & 84.2 & 85.4 & 86.4 & 70.7 \\
\textbf{monster} & 45.3 & 67.4 & 60.2 & 66.0 & 52.9 \\
\textbf{pickup} & 17.8 & 26.7 & 26.6 & 36.5 & 18.7 \\
semi & 83.9 & 79.8 & 79.0 & 83.5 & 86.1 \\
tank & 78.1 & 69.4 & 78.6 & 77.0 & 72.5 \\
\textbf{train} & 41.1 & 65.1 & 57.2 & 58.1 & 43.1 \\
van & 23.6 & 18.6 & 24.2 & 20.7 & 21.0 \\
\textbf{overall} & 52.3 & 59.2 & 59.2 & \textbf{61.2} & 52.3 \\
 \hline
\end{tabular}
\end{center}
\end{table}
\subsection{Class-agnostic Object Transformation Experiment}
Our proposed OPT-Net can simultaneously make pose transformation on different classes of vehicles, which demonstrate that the learned object pose transformation has not fixed with object classes, it is a class-agnostic object pose transformation. To further explore the class-agnostic property of OPT-Net, we design experiments that generalize OPT-Net’s ability for object pose transformation from one dataset to other datasets.

15 categories of objects from  RGB-D  are used. They are both common household objects with big spatial variance between different object poses. Similar poses of objects in RGB-D  are selected and defined as the same pose as iLab-20M. For each pose, RGB-D contains only about 100 images which cannot train our OPT-Net from scratch, thus we use RGB-D to finetune OPT-Net pre-trained on  iLab-20M. We can see (Fig. 5) that our pre-trained OPT-Net can generalize well to other datasets, which demonstrates that OPT-Net is a class-agnostic object pose transformation framework.

\begin{figure}[b]
\centering
\includegraphics[width=\linewidth]{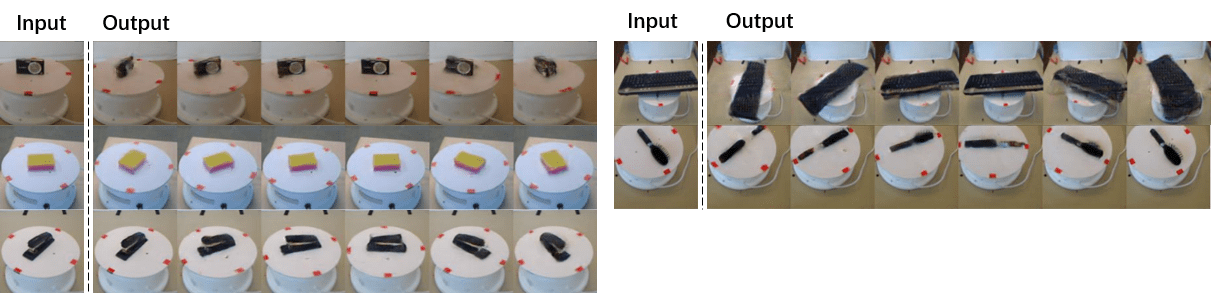}
\caption{Generalization results of OPT-Net on RGB-D dataset pretrained on iLab-20M.}
\end{figure}
To further explore the performance of OPT-Net as a generative model to help a discriminative model of object recognition, we split  RGB-D  into a pose-unbalanced (P-UB) training dataset, where each category randomly takes 3 poses among all 6 poses; pose-balanced (P-B), and test dataset similar to 4.2.

We first use P-UB to finetune the pretrained OPT-Net, and then use the trained OPT-Net to synthesize  missing poses of  household objects in RGB-D. The synthesized images and the original pose-unbalanced images form the synthesized pose balanced (S-P-B) training dataset. Similarly, to eliminate the influence of the number of training images, we created A-P-UB using common data augmentation methods. We trained Alexnet \cite{krizhevsky2012imagenet} on the 4 training datasets separately, and showed the test accuracy for each category in Table 5.

\begin{table}
\begin{center}
\caption{Overall object recognition accuracy for different training dataset in RGB-D}
\begin{tabular}{lllll}
\hline\noalign{\smallskip}
Dataset\ \ & P-UB\ \ \ & P-B\ \ \ \ \ & S-P-B\ \ \ & A-P-UB   \\
\noalign{\smallskip}
\hline
Accuracy(\%) & 99.1 & 99.9 & \textbf{99.7} & 99.2  \\
 \hline
\end{tabular}
\end{center}
\end{table}

The (small) accuracy improvement in S-P-B compared with P-UB demonstrates that our pretrained OPT-Net can be generalized to different datasets after finetune, which can help the discriminative model in object recognition. While the overall improvement is small, below we show that this is not the case uniformly across all object categories.

\subsection{Object Pose Significance on Different Object Recognition Tasks}
Because the accuracy improvement in RGB-D is smaller than  in  iLab-20M, we tested whether this was the case  across all object categories, or whether those which look more alike would benefit more from synthesized images from OPT-Net. Indeed, maybe classifying a black keyboard vs.\ a blue stapler can easily be achieved by size or color even without pose-dependent shape analysis.
To verify our hypothesis, we use the confusion matrix of classification to select categories which are more confused by classifier: marker, comb, toothbrush, stapler, lightbulb, and sponge.  We then assign different fixed poses to each category to improve overall pose variance and form P-UB-1 (randomly fix  1 pose for each category), P-UB-2 (randomly fix  2 poses for each category), and P-UB-3 (randomly fix  3 poses for each category) pose-unbalanced datasets (suppl. material).

Similarly, we create 3 other training datasets using the same method as in 4.2 and 4.3: (S-P-B: use pretrained OPT-Net to synthesize the missing poses; P-B, and A-P-UB for each unbalanced datasets), and report the object recognition performance on the test dataset in Table 6.

\begin{table}
\begin{center}
\caption{Object recognition overall accuracy for different datasets}
\begin{tabular}{lllllll}
\hline\noalign{\smallskip}
Dataset\ \ \ \ \ & P-UB-1\ \ \ & A-P-UB-1\ \ \ & S-P-B-1\ \ \ &P-UB-2\ \ \ & A-P-UB-2\ \ \ & S-P-B-2  \\
\noalign{\smallskip}
\hline
Accuracy(\%)  & 75.1 & 77.6 & \textbf{83.2} & 90.4 & 91.2 &\textbf{94.2}  \\
\hline\noalign{\smallskip}
Dataset  & P-UB-3 & A-P-UB-3 & S-P-B-3 &P-B   \\
\noalign{\smallskip}
\hline
Accuracy(\%)  & 99.3 & 99.2 & \textbf{99.4} &99.8  \\
\hline
\end{tabular}
\end{center}
\end{table}
The results in Table 6 demonstrate that object pose information has different degrees of impact on the object recognition task. Compared with the results in 4.3, where the improvement between P-UB and S-P-B is less than 1\%, here, when the class variance is small, OPT-Net can improve more accuracy after synthesizing the missing poses in the unbalanced dataset. The accuracy improvement in experiment group 1 (P-UB-1 and S-P-B-1) is 8.1\%. This result verified our hypothesis that pose balance is more important in small interclass variance object cognition tasks. Meanwhile, comparing the different accuracy improvements in different experimental groups, group 2 (P-UB-2 and S-P-B-2) is 3.8\%, while group 3 (P-UB-3 and S-P-B-3) is 0.1\%. This demonstrates that when class-variance is fixed, the more pose bias we have, the more accuracy improvement we will get with the help of our OPT-Net pose transformation.
\subsection{Generalization to Imagenet}
We directly use the pretrained OPT-Net on iLab-20M to synthesize images of different poses on ImageNet (Shown in suppl. material). Results are not as good and might be improved using domain adaptation in  future work. 
However, the discriminator of OPT-Net makes decent prediction of image poses: Fig. 6 shows the top 8 ImageNet images for each of our 6 poses.
\begin{figure}[t]
\centering
\includegraphics[height=3.1cm]{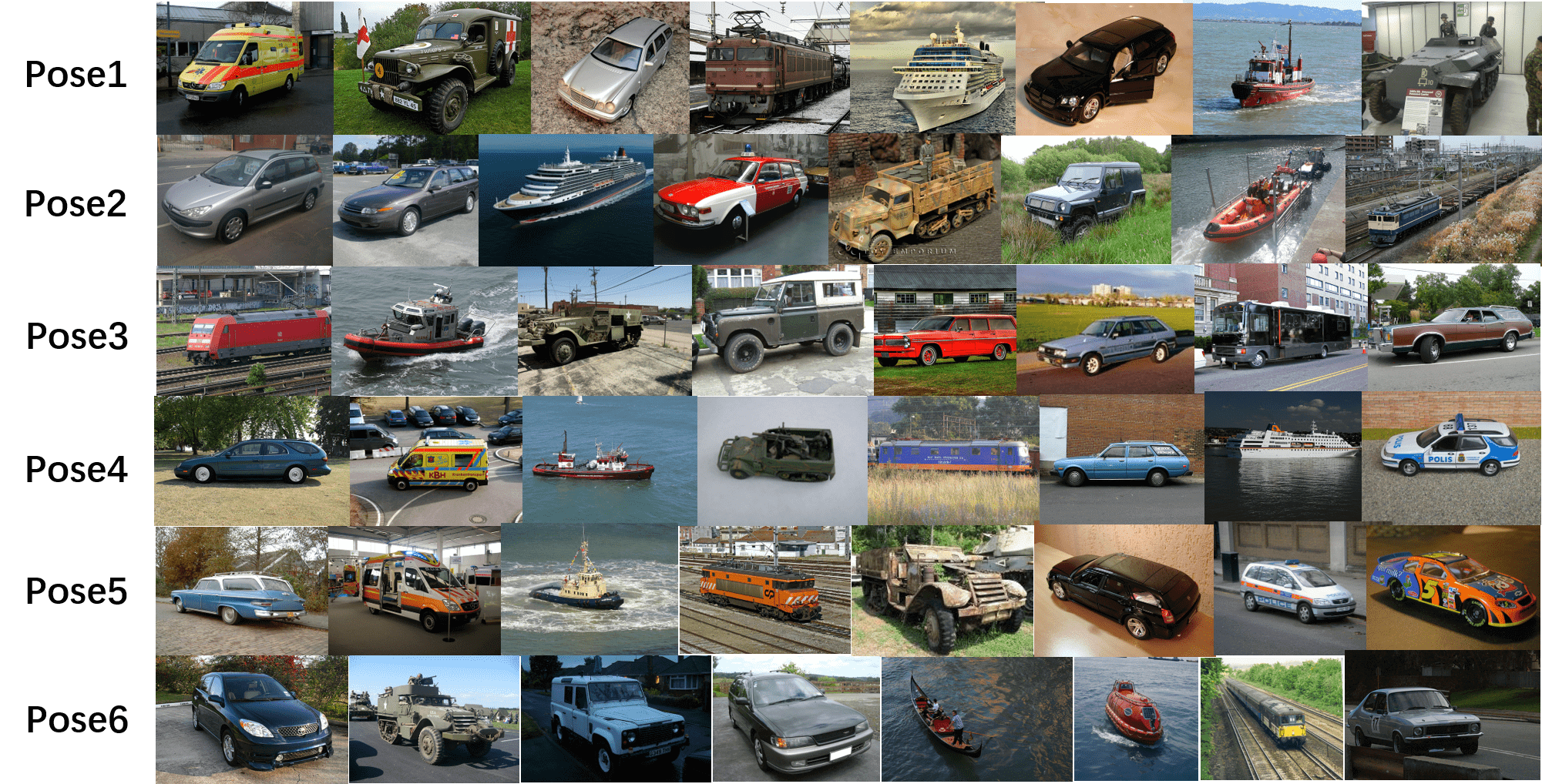}
\caption{Top 8 ImageNet images for each pose predicted by discriminator in OPT-Net without finetune.}
\end{figure}
To test object recognition in ImageNet, we replace real images by OPT-Net synthesized images in S-P-B (4.2) and form a S-P-B (OPT-Net) dataset (all synthesized images). Similarly, we use StarGAN synthesized images form S-P-B (StarGAN). We use a resnet18 10-class vehicles classifier pretrained with this two synthesized datasets and predict 4 classes of vehicles in ImageNet which have similar meanings as iLab-20M, with good results on some classes like car (Shown in suppl. material).
To further explore generalization, we pretrian an AlexNet on S-P-B which synthesized pose images by StarGAN and OPT-Net respectively and then finetune it on ImageNet. Results in suppl. material shows significantly better accuracy compared to training from scratch when using only a small number of images per class, demonstrating generalization from iLab-20M to ImageNet.

\section{Conclusions}
We proposed OPT-Net, a class-agnostic object pose transformation network (OPT-Net) to synthesize any target poses continuously given a single pose image. The proposed eliminate-add structure generator can first eliminate  pose information and turn the input to a pose-invariant canonical representation, then adding the target pose information to synthesize the target pose image. OPT-Net also gives a more common framework to solve big variance continuous transformation problems. OPT-Net generated images have higher visual quality compared to existing methods. We also demonstrate that the OPT-Net, as a generative model can help the discriminative model in the object recognition task, which achieve a 9\% accuracy improvement. We design experiments to demonstrate that pose balance is more important in small interclass variance object cognition tasks. Finally, we demonstrate the learned pose features in OPT-Net with the iLab-20M dataset can better generalize to other datasets like ImageNet.

{\bf Acknowledgements:}
This work was supported by C-BRIC (one of six centers in JUMP, a Semiconductor
Research Corporation (SRC) program sponsored by DARPA), and the Intel
and CISCO Corporations. The authors affirm that the views expressed
herein are solely their own, and do not represent the views of the
United States government or any agency thereof.

\clearpage
%
%


\begin{thebibliography}{10}
\providecommand{\url}[1]{\texttt{#1}}
\providecommand{\urlprefix}{URL }
\providecommand{\doi}[1]{https://doi.org/#1}

\bibitem{arjovsky2017wasserstein}
Arjovsky, M., Chintala, S., Bottou, L.: Wasserstein gan. arXiv preprint
  arXiv:1701.07875  (2017)

\bibitem{bakry2014untangling}
Bakry, A., Elgammal, A.: Untangling object-view manifold for multiview
  recognition and pose estimation. In: European Conference on Computer Vision.
  pp. 434--449. Springer (2014)

\bibitem{bengio2013representation}
Bengio, Y., Courville, A., Vincent, P.: Representation learning: A review and
  new perspectives. IEEE transactions on pattern analysis and machine
  intelligence  \textbf{35}(8),  1798--1828 (2013)

\bibitem{bhattacharjee2018posix}
Bhattacharjee, A., Banerjee, S., Das, S.: Posix-gan: Generating multiple poses
  using gan for pose-invariant face recognition. In: Proceedings of the
  European Conference on Computer Vision (ECCV). pp.~0--0 (2018)

\bibitem{borji2016ilab}
Borji, A., Izadi, S., Itti, L.: ilab-20m: A large-scale controlled object
  dataset to investigate deep learning. In: Proceedings of the IEEE Conference
  on Computer Vision and Pattern Recognition. pp. 2221--2230 (2016)

\bibitem{chang2018pairedcyclegan}
Chang, H., Lu, J., Yu, F., Finkelstein, A.: Pairedcyclegan: Asymmetric style
  transfer for applying and removing makeup. In: Proceedings of the IEEE
  Conference on Computer Vision and Pattern Recognition. pp. 40--48 (2018)

\bibitem{chen2016infogan}
Chen, X., Duan, Y., Houthooft, R., Schulman, J., Sutskever, I., Abbeel, P.:
  Infogan: Interpretable representation learning by information maximizing
  generative adversarial nets. In: Advances in neural information processing
  systems. pp. 2172--2180 (2016)

\bibitem{choi2018stargan}
Choi, Y., Choi, M., Kim, M., Ha, J.W., Kim, S., Choo, J.: Stargan: Unified
  generative adversarial networks for multi-domain image-to-image translation.
  In: Proceedings of the IEEE Conference on Computer Vision and Pattern
  Recognition. pp. 8789--8797 (2018)

\bibitem{deng2009imagenet}
Deng, J., Dong, W., Socher, R., Li, L.J., Li, K., Fei-Fei, L.: Imagenet: A
  large-scale hierarchical image database. In: 2009 IEEE conference on computer
  vision and pattern recognition. pp. 248--255. Ieee (2009)

\bibitem{fang2017rmpe}
Fang, H.S., Xie, S., Tai, Y.W., Lu, C.: Rmpe: Regional multi-person pose
  estimation. In: Proceedings of the IEEE International Conference on Computer
  Vision. pp. 2334--2343 (2017)

\bibitem{girshick2014rich}
Girshick, R., Donahue, J., Darrell, T., Malik, J.: Rich feature hierarchies for
  accurate object detection and semantic segmentation. In: Proceedings of the
  IEEE conference on computer vision and pattern recognition. pp. 580--587
  (2014)

\bibitem{goodfellow2014generative}
Goodfellow, I., Pouget-Abadie, J., Mirza, M., Xu, B., Warde-Farley, D., Ozair,
  S., Courville, A., Bengio, Y.: Generative adversarial nets. In: Advances in
  neural information processing systems. pp. 2672--2680 (2014)

\bibitem{goroshin2015learning}
Goroshin, R., Mathieu, M.F., LeCun, Y.: Learning to linearize under
  uncertainty. In: Advances in Neural Information Processing Systems. pp.
  1234--1242 (2015)

\bibitem{he2017mask}
He, K., Gkioxari, G., Doll{\'a}r, P., Girshick, R.: Mask r-cnn. In: Proceedings
  of the IEEE international conference on computer vision. pp. 2961--2969
  (2017)

\bibitem{he2016deep}
He, K., Zhang, X., Ren, S., Sun, J.: Deep residual learning for image
  recognition. In: Proceedings of the IEEE conference on computer vision and
  pattern recognition. pp. 770--778 (2016)

\bibitem{hoffman2017cycada}
Hoffman, J., Tzeng, E., Park, T., Zhu, J.Y., Isola, P., Saenko, K., Efros,
  A.A., Darrell, T.: Cycada: Cycle-consistent adversarial domain adaptation.
  arXiv preprint arXiv:1711.03213  (2017)

\bibitem{huang2017densely}
Huang, G., Liu, Z., Van Der~Maaten, L., Weinberger, K.Q.: Densely connected
  convolutional networks. In: Proceedings of the IEEE conference on computer
  vision and pattern recognition. pp. 4700--4708 (2017)

\bibitem{huang2013multi}
Huang, Y., Wang, W., Wang, L., Tan, T.: Multi-task deep neural network for
  multi-label learning. In: 2013 IEEE International Conference on Image
  Processing. pp. 2897--2900. IEEE (2013)

\bibitem{isola2017image}
Isola, P., Zhu, J.Y., Zhou, T., Efros, A.A.: Image-to-image translation with
  conditional adversarial networks. In: Proceedings of the IEEE conference on
  computer vision and pattern recognition. pp. 1125--1134 (2017)

\bibitem{kan2016multi}
Kan, M., Shan, S., Chen, X.: Multi-view deep network for cross-view
  classification. In: Proceedings of the IEEE Conference on Computer Vision and
  Pattern Recognition. pp. 4847--4855 (2016)

\bibitem{karras2019style}
Karras, T., Laine, S., Aila, T.: A style-based generator architecture for
  generative adversarial networks. In: Proceedings of the IEEE Conference on
  Computer Vision and Pattern Recognition. pp. 4401--4410 (2019)

\bibitem{kim2017learning}
Kim, T., Cha, M., Kim, H., Lee, J.K., Kim, J.: Learning to discover
  cross-domain relations with generative adversarial networks. In: Proceedings
  of the 34th International Conference on Machine Learning-Volume 70. pp.
  1857--1865. JMLR. org (2017)

\bibitem{krizhevsky2012imagenet}
Krizhevsky, A., Sutskever, I., Hinton, G.E.: Imagenet classification with deep
  convolutional neural networks. In: Advances in neural information processing
  systems. pp. 1097--1105 (2012)

\bibitem{lai2011large}
Lai, K., Bo, L., Ren, X., Fox, D.: A large-scale hierarchical multi-view rgb-d
  object dataset. In: 2011 IEEE international conference on robotics and
  automation. pp. 1817--1824. IEEE (2011)

\bibitem{lecun2015lenet}
LeCun, Y., et~al.: Lenet-5, convolutional neural networks

\bibitem{lee2019collagan}
Lee, D., Kim, J., Moon, W.J., Ye, J.C.: Collagan: Collaborative gan for missing
  image data imputation. In: Proceedings of the IEEE Conference on Computer
  Vision and Pattern Recognition. pp. 2487--2496 (2019)

\bibitem{ma2012invitation}
Ma, Y., Soatto, S., Kosecka, J., Sastry, S.S.: An invitation to 3-d vision:
  from images to geometric models, vol.~26. Springer Science \& Business Media
  (2012)

\bibitem{milletari2016v}
Milletari, F., Navab, N., Ahmadi, S.A.: V-net: Fully convolutional neural
  networks for volumetric medical image segmentation. In: 2016 Fourth
  International Conference on 3D Vision (3DV). pp. 565--571. IEEE (2016)

\bibitem{mirza2014conditional}
Mirza, M., Osindero, S.: Conditional generative adversarial nets. arXiv
  preprint arXiv:1411.1784  (2014)

\bibitem{ranzato2007unsupervised}
Ranzato, M., Huang, F.J., Boureau, Y.L., LeCun, Y.: Unsupervised learning of
  invariant feature hierarchies with applications to object recognition. In:
  2007 IEEE conference on computer vision and pattern recognition. pp.~1--8.
  IEEE (2007)

\bibitem{redmon2016you}
Redmon, J., Divvala, S., Girshick, R., Farhadi, A.: You only look once:
  Unified, real-time object detection. In: Proceedings of the IEEE conference
  on computer vision and pattern recognition. pp. 779--788 (2016)

\bibitem{ren2015faster}
Ren, S., He, K., Girshick, R., Sun, J.: Faster r-cnn: Towards real-time object
  detection with region proposal networks. In: Advances in neural information
  processing systems. pp. 91--99 (2015)

\bibitem{ronneberger2015u}
Ronneberger, O., Fischer, P., Brox, T.: U-net: Convolutional networks for
  biomedical image segmentation. In: International Conference on Medical image
  computing and computer-assisted intervention. pp. 234--241. Springer (2015)

\bibitem{sangkloy2017scribbler}
Sangkloy, P., Lu, J., Fang, C., Yu, F., Hays, J.: Scribbler: Controlling deep
  image synthesis with sketch and color. In: Proceedings of the IEEE Conference
  on Computer Vision and Pattern Recognition. pp. 5400--5409 (2017)

\bibitem{simonyan2014very}
Simonyan, K., Zisserman, A.: Very deep convolutional networks for large-scale
  image recognition. arXiv preprint arXiv:1409.1556  (2014)

\bibitem{su2015multi}
Su, H., Maji, S., Kalogerakis, E., Learned-Miller, E.: Multi-view convolutional
  neural networks for 3d shape recognition. In: Proceedings of the IEEE
  international conference on computer vision. pp. 945--953 (2015)

\bibitem{su2015render}
Su, H., Qi, C.R., Li, Y., Guibas, L.J.: Render for cnn: Viewpoint estimation in
  images using cnns trained with rendered 3d model views. In: Proceedings of
  the IEEE International Conference on Computer Vision. pp. 2686--2694 (2015)

\bibitem{szegedy2017inception}
Szegedy, C., Ioffe, S., Vanhoucke, V., Alemi, A.A.: Inception-v4,
  inception-resnet and the impact of residual connections on learning. In:
  Thirty-First AAAI Conference on Artificial Intelligence (2017)

\bibitem{tzeng2017adversarial}
Tzeng, E., Hoffman, J., Saenko, K., Darrell, T.: Adversarial discriminative
  domain adaptation. In: Proceedings of the IEEE Conference on Computer Vision
  and Pattern Recognition. pp. 7167--7176 (2017)

\bibitem{wohlhart2015learning}
Wohlhart, P., Lepetit, V.: Learning descriptors for object recognition and 3d
  pose estimation. In: Proceedings of the IEEE Conference on Computer Vision
  and Pattern Recognition. pp. 3109--3118 (2015)

\bibitem{zhang2014improving}
Zhang, C., Zhang, Z.: Improving multiview face detection with multi-task deep
  convolutional neural networks. In: IEEE Winter Conference on Applications of
  Computer Vision. pp. 1036--1041. IEEE (2014)

\bibitem{zhang2014facial}
Zhang, Z., Luo, P., Loy, C.C., Tang, X.: Facial landmark detection by deep
  multi-task learning. In: European conference on computer vision. pp. 94--108.
  Springer (2014)

\bibitem{zhao2017learning}
Zhao, J., Chang, C.k., Itti, L.: Learning to recognize objects by retaining
  other factors of variation. In: 2017 IEEE Winter Conference on Applications
  of Computer Vision (WACV). pp. 560--568. IEEE (2017)

\bibitem{zhao2015stacked}
Zhao, J., Mathieu, M., Goroshin, R., Lecun, Y.: Stacked what-where
  auto-encoders. arXiv preprint arXiv:1506.02351  (2015)

\bibitem{zhu2017unpaired}
Zhu, J.Y., Park, T., Isola, P., Efros, A.A.: Unpaired image-to-image
  translation using cycle-consistent adversarial networks. In: Proceedings of
  the IEEE international conference on computer vision. pp. 2223--2232 (2017)

\bibitem{zhu2014multi}
Zhu, Z., Luo, P., Wang, X., Tang, X.: Multi-view perceptron: a deep model for
  learning face identity and view representations. In: Advances in Neural
  Information Processing Systems. pp. 217--225 (2014)

\end{thebibliography}

\clearpage

\section{Supplementary Materials}
\subsection{Network structure and Training details}
We mentioned the high level network architecture and training instructions in main paper Section 3.2. Here are the details.

\textbf{Network structure} OPT-Net consists of 3 modules: eliminate-add structure generator $G$ (which consists of $G_{elim}$ and $G_{add}$) (shown in Table 1), discriminator $D$ (shown in Table 2), and pose-eliminate module $P_{elim}$ (shown in Table 3).  Here are some notations; $n_{p}$: number of predefined discrete poses; $n_{3D}$: channel dimension of added pose information mask representing yaw and pitch values; Conv-(): convolutional layer; DeConv-(): deconvolutional layer, $n$: number of convolution kernels, $k$: kernel size, $s$: stride size, $p$: padding size, In: instance normalization, Res: Residual Block.

\begin{table}[h]
\scriptsize
\centering
\caption{Eliminate-add structure generator $G$ network architecture}
\begin{tabular}{ccc}
\hline\noalign{\smallskip}
Part & Input $\rightarrow$ Output Shape & Layer Information\\
\noalign{\smallskip}
 \hline
 \hline
 \noalign{\smallskip}
 \multirow{6}*{$G_{elim}$} & (h, w, 3) $\rightarrow$ (h, w, 64) & Conv-($n$64, $k$7x7, $s$1, $p$3), In, Relu\\\noalign{\smallskip}
 ~ & \ \  (h, w, 64) $\rightarrow$ ($\frac{h}{2}$,  $\frac{w}{2}$, 128) & Conv-($n$128, $k$4x4, $s$2, $p$1), In, Relu \\\noalign{\smallskip}
 ~ & \ \  ($\frac{h}{2}$,  $\frac{w}{2}$, 128) $\rightarrow$ ($\frac{h}{4}$,  $\frac{w}{4}$, 256) & Conv-($n$256, $k$4x4, $s$2, $p$1), In, Relu\\\noalign{\smallskip}
 ~ & \ \  ($\frac{h}{4}$,  $\frac{w}{4}$, 256)$\rightarrow$($\frac{h}{4}$,  $\frac{w}{4}$, 256) & \ \ \ Res: Conv-($n$256, $k$3x3, $s$1, $p$1), In, Relu \\\noalign{\smallskip}
 ~ & \ \  ($\frac{h}{4}$,  $\frac{w}{4}$, 256)$\rightarrow$($\frac{h}{4}$,  $\frac{w}{4}$, 256) & \ \ \ Res: Conv-($n$256, $k$3x3, $s$1, $p$1), In, Relu \\\noalign{\smallskip}
 ~ & \ \  ($\frac{h}{4}$,  $\frac{w}{4}$, 256)$\rightarrow$($\frac{h}{4}$,  $\frac{w}{4}$, 256) & \ \ \ Res: Conv-($n$256, $k$3x3, $s$1, $p$1), In, Relu \\\noalign{\smallskip}
 \noalign{\smallskip}
 \hline
 \noalign{\smallskip}
 \multirow{9}*{$G_{add}$}  & \ \  ( $\frac{h}{4}$,  $\frac{w}{4}$, 256 + $\bf{n_{3D}}$ )   $\rightarrow$  ($\frac{h}{4}$,  $\frac{w}{4}$, 256) & \ \ \  Conv-($n$256, $k$3x3, $s$1, $p$1), In, Relu \\\noalign{\smallskip}
 ~ & \ \  ($\frac{h}{4}$,  $\frac{w}{4}$, 256) $\rightarrow$ ($\frac{h}{4}$,  $\frac{w}{4}$, 256) & \ \ \ Res: Conv-($n$256, $k$3x3, $s$1, $p$1), In, Relu \\\noalign{\smallskip}
 ~ & \ \  ($\frac{h}{4}$,  $\frac{w}{4}$, 256) $\rightarrow$ ($\frac{h}{4}$,  $\frac{w}{4}$, 256) & \ \ \ Res: Conv-($n$256, $k$3x3, $s$1, $p$1), In, Relu \\\noalign{\smallskip}
  ~ & \ \  ($\frac{h}{4}$,  $\frac{w}{4}$, 256) $\rightarrow$ ($\frac{h}{4}$,  $\frac{w}{4}$, 256) & \ \ \ Res: Conv-($n$256, $k$3x3, $s$1, $p$1), In, Relu \\\noalign{\smallskip}
 ~ & \ \  ($\frac{h}{4}$,  $\frac{w}{4}$, 256) $\rightarrow$ ($\frac{h}{4}$,  $\frac{w}{4}$, 256) & \ \ \ Res: Conv-($n$256, $k$3x3, $s$1, $p$1), In, Relu \\\noalign{\smallskip}
  ~ & \ \  ($\frac{h}{4}$,  $\frac{w}{4}$, 256) $\rightarrow$ ($\frac{h}{4}$,  $\frac{w}{4}$, 256) & \ \ \ Res: Conv-($n$256, $k$3x3, $s$1, $p$1), In, Relu \\\noalign{\smallskip}
  
 ~ &($\frac{h}{4}$,  $\frac{w}{4}$, 256)$\rightarrow$($\frac{h}{2}$,  $\frac{w}{2}$, 128) & DeConv-($n$128, $k$4x4, $s$2, $p$1), In, Relu\\\noalign{\smallskip}
 ~ & \ \  $\frac{h}{2}$,  $\frac{w}{2}$, 128)$\rightarrow$(h, w, 64) & DeConv-($n$64, $k$4x4, $s$2, $p$1), In, Relu \\\noalign{\smallskip}
 ~ & \ \  (h, w, 64)$\rightarrow$(h, w, 3) & Conv-($n$3, $k$7x7, $s$1, $p$3), In, Tanh\\
 \noalign{\smallskip}
\hline
 
\end{tabular}
\end{table}

\begin{table}[h]
\scriptsize
\centering
\caption{Discriminator $D$ network architecture}
\begin{tabular}{ccc}
\hline\noalign{\smallskip}
Layer & Input $\rightarrow$ Output Shape & Layer Information\\
\noalign{\smallskip}
 \hline
 \hline
 \noalign{\smallskip}
 Input Layer & (h, w, 3) $\rightarrow$ ($\frac{h}{2}$,  $\frac{w}{2}$, 64) & Conv-($n$64, $k$4x4, $s$2, $p$1), Leaky ReLU \\\noalign{\smallskip}
 Hidden Layer & \ \  ($\frac{h}{2}$,  $\frac{w}{2}$, 64) $\rightarrow$ ($\frac{h}{4}$,  $\frac{w}{4}$, 128) & Conv-($n$128, $k$4x4, $s$2, $p$1), Leaky ReLU \\\noalign{\smallskip}
 Hidden Layer & \ \  ($\frac{h}{4}$,  $\frac{w}{4}$, 128) $\rightarrow$ ($\frac{h}{8}$,  $\frac{w}{8}$, 256) & Conv-($n$256, $k$4x4, $s$2, $p$1),  Leaky ReLU\\\noalign{\smallskip}
 Hidden Layer & \ \  ($\frac{h}{8}$,  $\frac{w}{8}$, 256)$\rightarrow$($\frac{h}{16}$,  $\frac{w}{16}$, 512) & \ \ \ Conv-(512, $k$4x4, $s$2, $p$1), Leaky ReLU \\\noalign{\smallskip}
 Hidden Layer & \ \  ($\frac{h}{16}$,  $\frac{w}{16}$, 512)$\rightarrow$($\frac{h}{32}$,  $\frac{w}{32}$, 1024) & \ \ \ Conv-(1024, $k$4x4, $s$2, $p$1), Leaky ReLU \\\noalign{\smallskip}
 Hidden Layer & \ \  ($\frac{h}{32}$,  $\frac{w}{32}$, 1024)$\rightarrow$($\frac{h}{64}$,  $\frac{w}{64}$, 2048) & \ \ \ Conv-($n$2048, $k$4x4, $s$2, $p$1), Leaky ReLU \\\noalign{\smallskip}
\noalign{\smallskip}
\hline
\noalign{\smallskip}
 Output Layer($ D_{src}$) & \ \  ($\frac{h}{64}$,  $\frac{w}{64}$, 2048)$\rightarrow$($\frac{h}{64}$,  $\frac{w}{64}$, 1) & \ \ \ Conv-($n$1, $k$3x3, $s$1, $p$1) \\\noalign{\smallskip}
 Output Layer($ D_{cls}$) & \ \  ($\frac{h}{64}$,  $\frac{w}{64}$, 2048)$\rightarrow$(1,  1, $n_{p}$) & \ \ \ Conv-($n$($n_{p}$), $k$ $\frac{h}{64}$ x $\frac{h}{64}$, $s$1, $p$0) \\\noalign{\smallskip}
 \hline
\end{tabular}
\end{table}

\begin{table}[h]
\scriptsize
\centering
\caption{Pose-eliminate module $P_{elim}$ network architecture}
\begin{tabular}{ccc}
\hline\noalign{\smallskip}
Layer & Input $\rightarrow$ Output Shape & Layer Information\\
\noalign{\smallskip}
 \hline
 \hline
 \noalign{\smallskip}
 Input Layer & (h, w, 3) $\rightarrow$ ($\frac{h}{2}$,  $\frac{w}{2}$, 64) & Conv-($n$64, $k$4x4, $s$2, $p$1), Leaky ReLU \\\noalign{\smallskip}
 Hidden Layer & \ \  ($\frac{h}{2}$,  $\frac{w}{2}$, 64) $\rightarrow$ ($\frac{h}{4}$,  $\frac{w}{4}$, 128) & Conv-($n$128, $k$4x4, $s$2, $p$1), Leaky ReLU \\\noalign{\smallskip}
 Hidden Layer & \ \  ($\frac{h}{4}$,  $\frac{w}{4}$, 128) $\rightarrow$ ($\frac{h}{8}$,  $\frac{w}{8}$, 256) & Conv-($n$256, $k$4x4, $s$2, $p$1),  Leaky ReLU\\\noalign{\smallskip}
 Residual Block & \ \  ($\frac{h}{8}$,  $\frac{w}{8}$, 256)$\rightarrow$($\frac{h}{8}$,  $\frac{w}{8}$, 256) & \ \ \ Res: Conv-(256, $k$3x3, $s$1, $p$1), Leaky ReLU \\\noalign{\smallskip}

 Hidden Layer & \ \  ($\frac{h}{8}$,  $\frac{w}{8}$, 256)$\rightarrow$($\frac{h}{16}$,  $\frac{w}{16}$, 512) & \ \ \ Conv-(512, $k$4x4, $s$2, $p$1), Leaky ReLU \\\noalign{\smallskip}
 Hidden Layer & \ \  ($\frac{h}{16}$,  $\frac{w}{16}$, 512)$\rightarrow$($\frac{h}{32}$,  $\frac{w}{32}$, 1024) & \ \ \ Conv-(1024, $k$4x4, $s$2, $p$1), Leaky ReLU \\\noalign{\smallskip}
 Hidden Layer & \ \  ($\frac{h}{32}$,  $\frac{w}{32}$, 1024)$\rightarrow$($\frac{h}{64}$,  $\frac{w}{64}$, 2048) & \ \ \ Conv-($n$2048, $k$4x4, $s$2, $p$1), Leaky ReLU \\\noalign{\smallskip}
\noalign{\smallskip}
\hline
\noalign{\smallskip}
 Output Layer & \ \  ($\frac{h}{64}$,  $\frac{w}{64}$, 2048)$\rightarrow$(1,  1, $n_{p}$) & \ \ \ Conv-($n$($n_{p}$), $k$ $\frac{h}{64}$ x $\frac{h}{64}$, $s$1, $p$0) \\\noalign{\smallskip}
 \hline
\end{tabular}
\end{table}

\textbf{Training details} We train OPT-Net on iLab-20M (P-UB) dataset using Adam with $\beta_{1}$=0.5 and $\beta_{2}$=0.999, batch size 64, learning rate 0.0001 for the first 500 epochs and linear decay to 0 over the next 500 epochs.

Hyperparameters in the loss function: For Pose classification loss and Reconstruction loss, we use $\lambda_{cls}$ = 1, $\lambda_{rec}$ = 10 unmodified. For Pose-eliminate loss, as we described at the end of main paper 3.2, we set $\lambda_{pose}$ = 2 at first 1/4 of all epochs (250 epochs), and then we linearly decay the $\lambda_{pose}$ to 0 over the next 750 epochs.

During training, when we add new loss terms to GAN structure, we need to monitor the adversarial loss and make sure the new term will not break the balance. Moreover, for some loss to refine the quality or perform as regularization, like style consistency, we can monitor the performance of synthesized images and add them after the generator can synthesize good quality images.

\subsection{Object pose significance on different object recognition tasks}
As we described in main paper 4.4, after selecting categories that are more confused by the classifier: marker, comb, toothbrush, stapler, lightbulb, and sponge, we assign different fixed poses for each category to improve overall pose variance and form P-UB-1, P-UB-2, and P-UB-3 pose-unbalanced datasets. Table 4 shows the pose category details in different unbalanced datasets. 

\begin{table}
\begin{center}
\begin{tabular}{llll}
\hline
category \ \ \ \  & Pose in P-UB-1 \ \ \ \   & Pose in P-UB-2 \ \ \ \   & Pose in P-UB-3 \\
\hline
comb & 1 & 1,2 & 1,2,3  \\
 \hline
 toothpaste & 2 & 2,3 & 2,3,4  \\
 \hline
 flashlight & 3 & 3,4 & 3,4,5  \\
 \hline
 lightbulb & 4 & 4,5 & 4,5,6  \\
 \hline
 marker & 5 & 5,6 & 1,5,6  \\
 \hline
 sponge & 6 & 1,6 & 1,2,6  \\
 \hline

\end{tabular}
\end{center}
\caption{Available pose categories in different unbalanced datasets}
\end{table}
\subsection{Generalization to Imagenet}

\textbf{ImageNet image pose transform without finetune} As we described in main paper 4.5, we directly use the pretrained OPT-Net on iLab-20M to synthesize different pose images for images from ImageNet without finetuning (Fig.1).

\begin{figure}
\centering
\includegraphics[width=\linewidth]{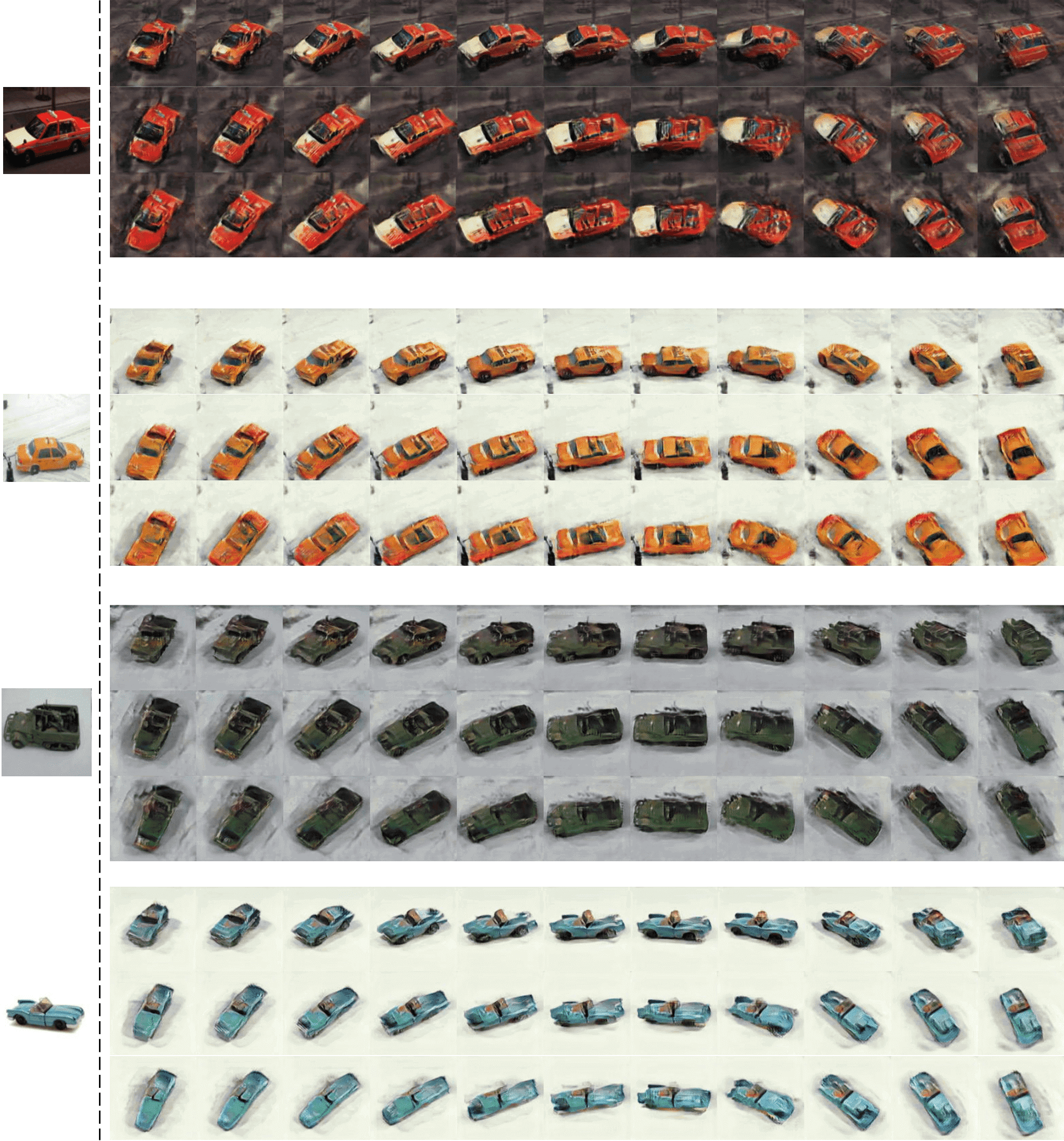}
\caption{Directly synthesized different-posed ImageNet images by OPT-Net without finetune.The first column shows input images from ImageNet, and the remaining columns show target pose images transformed by OPT-Net.  Each row demonstrates the pose change along yaw axis, each column shows pose changes along pitch axis.}
\end{figure}
Although the general shape and pose structure are reasonable, the image quality is not as good as the synthesized images in iLab-20M. We interpret this result for two reasons. First, the object pose, background, and camera view of images in ImageNet are different from that in iLab-20M, which may cause domain gap and inferior quality. Second, we do not finetune OPT-Net on ImageNet, which may degrade the generation quality as well. It might be improved using domain adaptation or fine tuning in future work.

\textbf{OPT-Net performance on boost object recognition in ImageNet} As we described in main paper 4.5, we replace real images by OPT-Net synthesized images in S-P-B (main paper 4.2) and form a S-P-B (OPT-Net) dataset (all synthesized images). Similarly, we use StarGAN synthesized images form S-P-B (StarGAN). We use a resnet18 10-class vehicles classifier pretrained with this two synthesized datasets and predict 4 classes of vehicles in ImageNet which have similar meanings as iLab-20M, with good results on some classes like car (Shown in Table 5.).

Table 5 demonstrate two main points, first, a pre-trained OPT-Net learns some knowledge from iLab-20M, which can be directly used to boost classification tasks on ImageNet with no finetune (by synthesizing new images). Second, the synthesized performance of OPT-Net is better than other methods.

\begin{table}[t]
\begin{center}
\caption{Generalization to ImageNet (percent correct)}
\begin{tabular}{lllll}
\hline\noalign{\smallskip}
Pretrain dataset\ \ \ \ \ \ \ & boat\ \ \ \ \ & car\ \ \ \ \ & mil\ \ \ \ \ & tank \\
\noalign{\smallskip}
\hline
No pretrain & 0 & 0 & 0 & 0  \\
S-P-B (StarGAN) & 16.33 & 41.84 & 17.20 & 6.21  \\
S-P-B (OPT-Net) & \textbf{31.14} & \textbf{76.20} & \textbf{30.47} & \textbf{12.86}  \\
\hline
\end{tabular}
\end{center}
\end{table}

\textbf{Generalization from iLab-20M to ImageNet} As we described in main paper 4.5, to further explore generalization, we pretrian an AlexNet on S-P-B which synthesized pose images by StarGAN and OPT-Net respectively and then finetune it on ImageNet with limited images per class (randomly choose 10, 20, and 40 images for each class). Results in Table 6 show better accuracy compared to training from scratch when using only a small number of images per class, demonstrating generalization from iLab-20M to ImageNet.

\begin{table}
\begin{center}
\caption{Top-5 recognition accuracy (\%) on ILSVRC-2010 test set}
\begin{tabular}{llll}
\hline\noalign{\smallskip}
Number of images per class \ \ \ \ \ \ \ \  \ \ \ \ \ \  \ \ \ \ \ \ \ \ \ \ & 10 & 20 & 40  \\
\noalign{\smallskip}
\hline
Alexnet (scratch)\ \ \ \ & 4.32\ \ \ & 17.01\ \ \ & 25.35  \\
Alexnet (pretrain on S-P-B (StarGAN)) & 7.57 & 18.91 & 27.14  \\
Alexnet (pretrain on S-P-B (OPT-Net)) & \textbf{10.94} & \textbf{19.75} & \textbf{27.82}  \\
\hline
\end{tabular}
\end{center}
\end{table}

\subsection{Additional Qualitative Pose Transformation Results of OPT-Net}

Figs. 2, 3, and 4 show additional images with 128 $\times$ 128 resolutions pose transformation results generated by OPT-Net. All images were generated by a eliminate-add structure generator trained on iLab-20M datasets. The first column shows input images from the test dataset, and the remaining columns show target pose images transformed by OPT-Net.  Poses framed in red are defined in the training dataset, while all other poses are new poses, which shows OPT-Net can achieve continuous pose transformation.
\begin{figure}
\centering
\includegraphics[width=\linewidth]{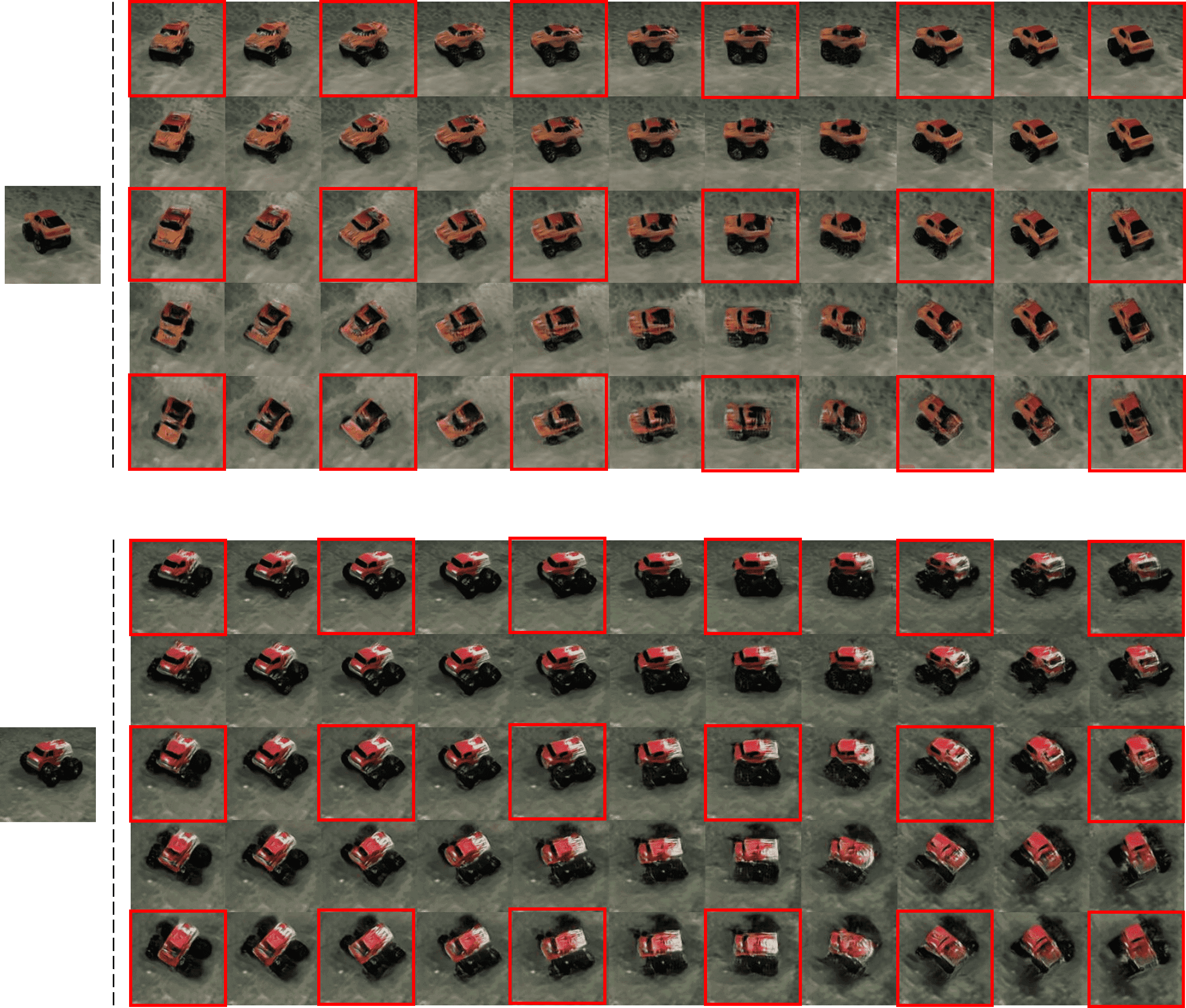}
\caption{OPT-Net pose transformation results on monster vehicle, The first column shows input images from the test dataset, and the remaining columns show target pose images transformed by OPT-Net.  Poses framed in red are defined in the training dataset, while all other poses are new poses, which shows OPT-Net can achieve continuous pose transformation}
\end{figure}

\begin{figure}
\centering
\includegraphics[width=\linewidth]{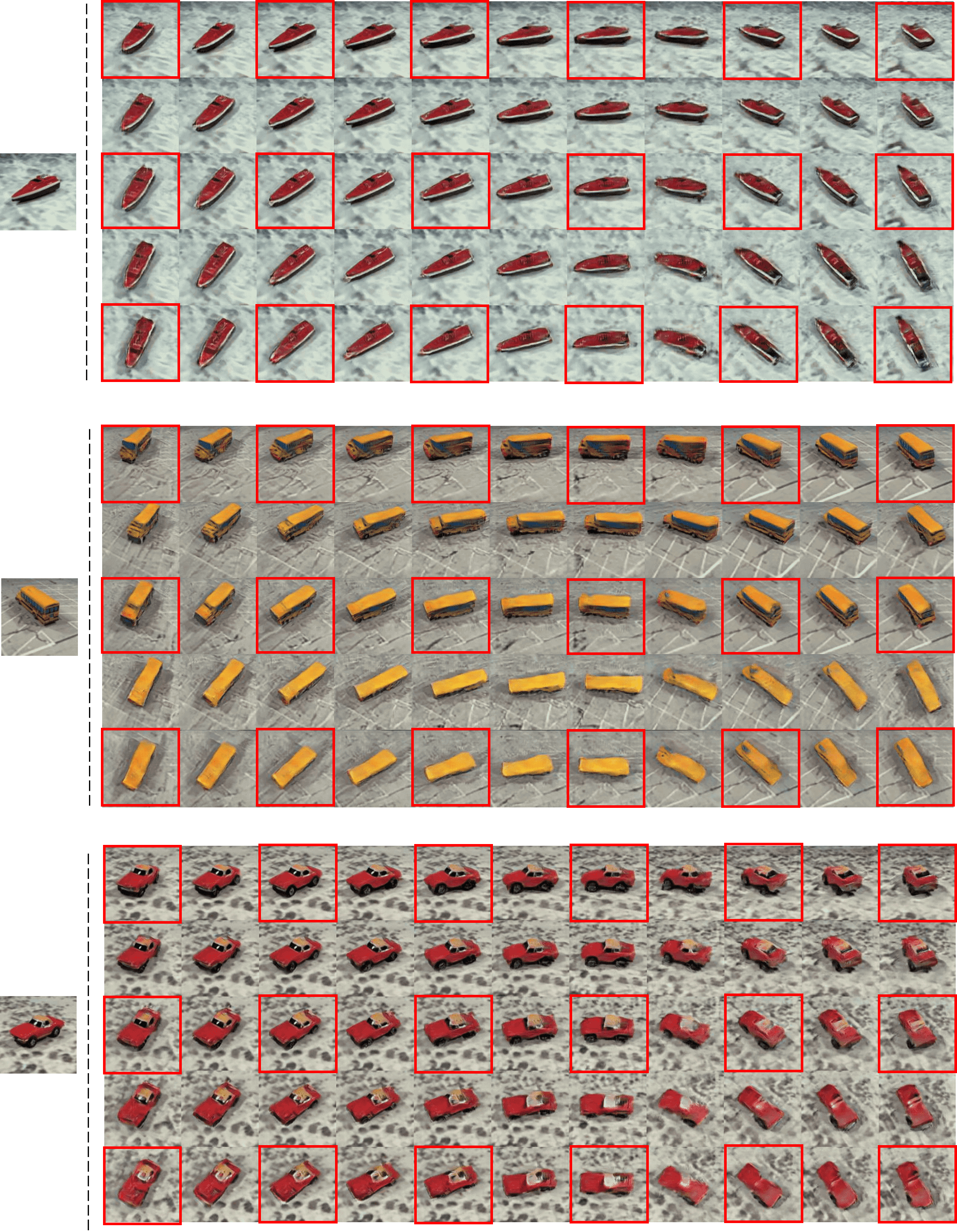}
\caption{OPT-Net pose transformation results on boat, bus and car, The first column shows input images from the test dataset, and the remaining columns show target pose images transformed by OPT-Net.  Poses framed in red are defined in the training dataset, while all other poses are new poses, which shows OPT-Net can achieve continuous pose transformation}
\end{figure}

\begin{figure}
\centering
\includegraphics[width=\linewidth]{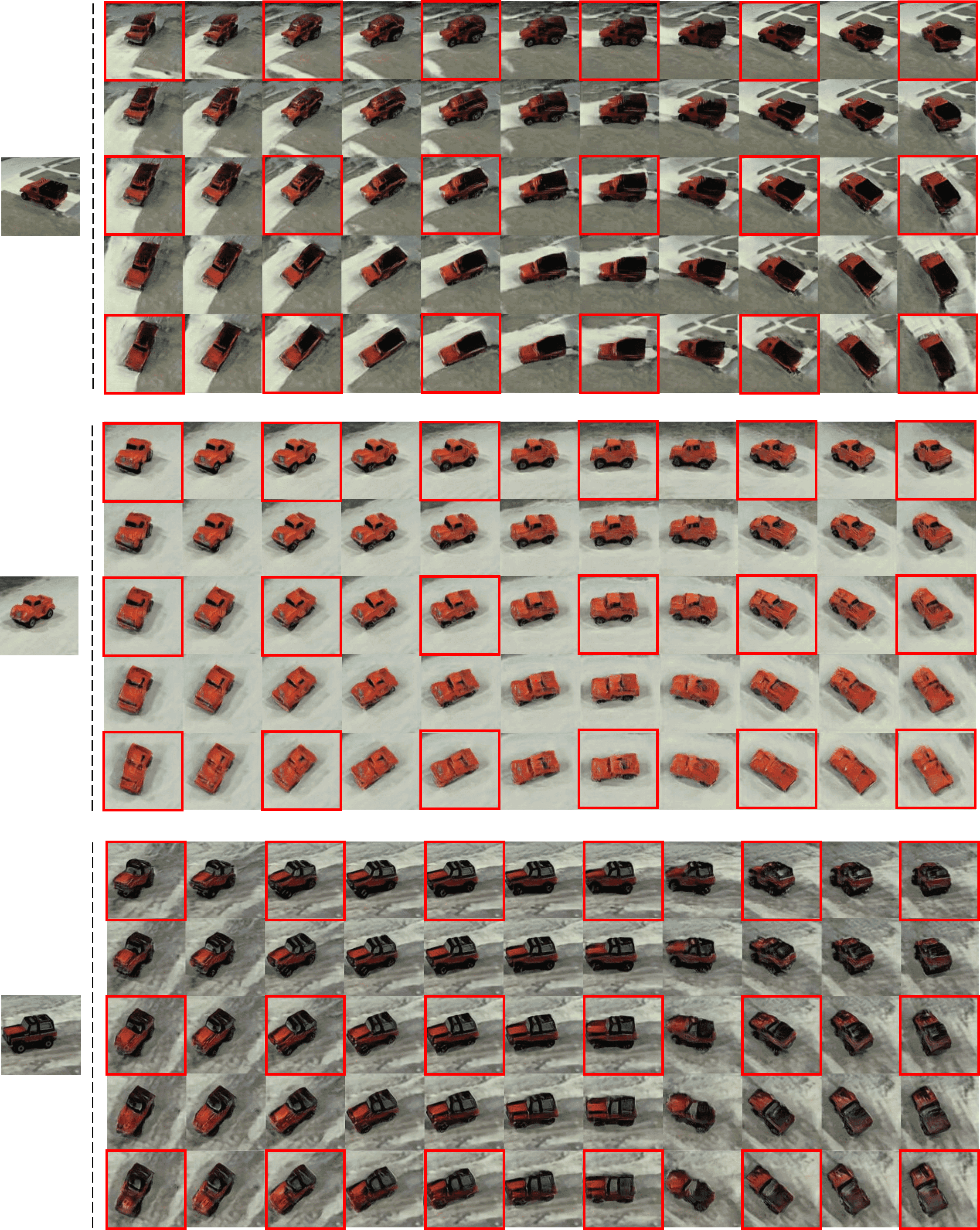}
\caption{OPT-Net pose transformation results on pick-up vehicle, The first column shows input images from the test dataset, and the remaining columns show target pose images transformed by OPT-Net.  Poses framed in red are defined in the training dataset, while all other poses are new poses, which shows OPT-Net can achieve continuous pose transformation}
\end{figure}

\end{document}